\documentclass[12pt]{article}
\usepackage{amsmath}
\usepackage{times}
\usepackage{graphicx}
\usepackage{color}
\usepackage{multirow}
\usepackage[authoryear]{natbib}
\usepackage{rotating}
\usepackage{bbm}
\usepackage{latexsym}
\usepackage{amssymb}

\newcommand{\captionfonts}{\normalsize}

\makeatletter  
\long\def\@makecaption#1#2{%
  \vskip\abovecaptionskip
  \sbox\@tempboxa{{\captionfonts #1: #2}}%
  \ifdim \wd\@tempboxa >\hsize
    {\captionfonts #1: #2\par}
  \else
    \hbox to\hsize{\hfil\box\@tempboxa\hfil}%
  \fi
  \vskip\belowcaptionskip}
\makeatother   

\begin{document}
\hspace{13.9cm}1

\ \vspace{20mm}\\


{\LARGE Sensitivity of sparse codes to image distortions}

\ \\
{\bf \large Kyle Luther$^{\displaystyle 1, \displaystyle 2}$}\\
{\bf \large H. Sebastian Seung$^{\displaystyle 2, \displaystyle 3}$}\\
{$^{\displaystyle 1}$Department of Physics, Princeton University, Princeton, New Jersey 08544, USA}\\
{$^{\displaystyle 2}$Neuroscience Institute, Princeton University, Princeton, New Jersey 08544, USA}\\
{$^{\displaystyle 3}$Department of Computer Science, Princeton University, Princeton, New Jersey 08544, USA}\\


\thispagestyle{empty}
\markboth{}{NC instructions}
\ \vspace{-0mm}\\
%
\begin{abstract}
\normalsize
Sparse coding has been proposed as a theory of visual cortex and as an unsupervised algorithm for learning representations. We show empirically with the MNIST dataset that sparse codes can be very sensitive to image distortions, a behavior that may hinder invariant object recognition. A locally linear analysis suggests that the sensitivity is due to the existence of linear combinations of active dictionary elements with high cancellation. A nearest neighbor classifier is shown to perform worse on sparse codes than original images. For a linear classifier with a sufficiently large number of labeled examples, sparse codes are shown to yield higher accuracy than original images, but no higher than a representation computed by a random feedforward net. Sensitivity to distortions seems to be a basic property of sparse codes, and one should be aware of this property when applying sparse codes to invariant object recognition.
\end{abstract}

\section{Introduction}

Given a generative model $f$ for images, one can encode an image $\mathbf{x}$ by a representation vector $\mathbf{r}$ such that $\mathbf{x} \approx f(\mathbf{r})$. While generative models have many possible uses, this paper is concerned with their application to visual object recognition. In this approach, a classifier is trained to predict the object class based on the representation vector rather than the image. It is hoped that the approach will reduce the number of labels required to train the classifier, or reduce the complexity of the classifier.

The approach has a rich history. It has been tried with variational autoencoders \citep{kingma2014semisupervised}, generative adversarial networks \citep{radford2016gan}, denoising autoencoders \citep{vincent2010stacked}, sparse coding \citep{raina2007selftaught}, non-negative matrix factorization \citep{guillamet2002nmfclassification}, and principal components analysis \citep{turk1991eigenfaces}. However, the approach has largely been abandoned in favor of purely supervised training of classifiers that use the image as input. More recently, progress has been made with a contrastive approach to representation learning in which network output is trained to be invariant to a hand-designed set of augmentations \citep{dosovitskiy2014discriminative,chen2020simclr,he2020moco,caron2020swav,deny2021barlowtwins}. 

Nevertheless, generative models remain appealing because neither labels nor a hand-designed set of invariances are needed to train them. The goal of this paper is to identify and study an issue that might be hindering the success of the generative model approach to object recognition. For concreteness, we focus our attention on sparse linear generative models, otherwise known as \textit{sparse coding} \citep{olshausen1996emergence}. The goal of sparse coding is to learn a dictionary $\mathbf{D} \in \mathbb{R}^{m\times n}$ and for every image $\mathbf{x} \in \mathbb{R}^m$ a sparse representation $\mathbf{r} \in \mathbb{R}^n$ such that $\mathbf{x}$ is approximated by a linear combination of a small subset of dictionary elements: $\mathbf{x} \approx \mathbf{D}\mathbf{r}$.

Sparse coding has been proposed as a theory of visual cortex, because the representations can be computed by a recurrent neural network in which neurons inhibit each other through lateral interactions, the receptive fields of neurons end up resembling Gabor filters, and the representations are sparse \citep{olshausen1997overcomplete}. 


The goal of our paper is to point out sparse codes can be very sensitive to image distortions, a property that may ultimately be at odds with the task of object recognition. We demonstrate this empirically using the MNIST images of handwritten digits. Sparse codes turn out to be more sensitive to an image distortion that preserves the digit class than to an image perturbation that can change the digit class. Furthermore, this sensitivity emerges in the sparse codes; it is not present in the original images.

To explain these empirical findings, we use the locally linear relationship between the sparse code and the image, which is based on the dictionary elements or filters of the sparse code. Sensitivity is related to the existence of linear combinations of active filters with high cancellation. These define directions in image space that turn out to be well-aligned with image distortions for MNIST, leading to high sensitivity.

We examine the impact of sensitivity to distortions on downstream classification performance. Nearest neighbor classifiers based on sparse codes significantly underperform those based on raw image pixels. For a linear classifier with a sufficiently large number of labeled examples, we find that sparse codes can yield higher accuracy than raw image pixels, consistent with previous work \citep{raina2007selftaught}. However, the increased linear classifier accuracy is matched by a representation computed by a random feedforward net \citep{jarrett2009best}. For a linear classifier with few labeled examples, sparse codes are outperformed by both the raw image pixels and the random net representation. 

Our results seem to suggest that sparse coding is useful for downstream recognition only insofar as it increases linear separability, but also introduces a high sensitivity to distortions which may ultimately be at odds with the task of invariant object recognition.

Some researchers do not take note of the sensitivity issue or do not regard it as problematic \citep{raina2007selftaught}. We have also encountered colleagues who believe that the Restricted Isometry Property (RIP) from compressed sensing \citep{candes2008restricted} guarantees that there is no sensitivity issue. In fact, there is no reason to expect that the RIP should apply to sparse coding, and we will empirically demonstrate RIP violation in the main text.

Other researchers have introduced postprocessing mechanisms to reduce the sensitivity of sparse codes. Spatial pooling has been applied to convolutional sparse codes \citep{hu2014sparsity, yang2009linear}. \citet{chen2018smt} used temporal information to learn a linear pooling operator, similar in spirit to applying slow feature analysis to sparse codes, with the goal of linearizing representation trajectories over time. Other work has modified the original formulation of sparse coding to use complex basis functions, where the phase encodes position and the amplitude encodes a local invariance \citep{cadieu2012learning}.

These previous works can be viewed as attempts to reduce the sensitivity behavior. Here we provide an explicit formulation of the behavior: the sparse codes themselves may be more sensitive than the raw pixels to small input distortions. We hope that it will be a helpful starting point for other researchers who hope to make sparse coding competitive with purely supervised approaches to object recognition, or with the popular contrastive approach to self-supervised representation learning \citep{chen2020simclr, caron2020swav, grill2020byol}.

We also note that the contrastive approach works by training a net to be insensitive to image distortions that preserve the object class. In this sense, our formulation of the sensitivity behavior views sparse coding through a lens provided by the contrastive approach to representation learning.

\section{Sparse codes are sensitive to image distortions}

In this section, we train sparse coding on the MNIST dataset \citep{deng2012mnist} and measure the sensitivity of the generated representations, defined by the norm of their directional derivative, to image perturbations shown in Figure \ref{fig:input-perturbation}. The motivation for this experiment is to measure the invariance properties of the generated representations. The fundamental challenge of visual object recognition is knowing which of the enormous number of pixel variations are irrelevant for discrimination and which are relevant for discrimination \citep{dicarlo2012brain}. If the goal of representation learning is ultimately to generate representations which are useful for downstream recognition, it seems natural to expect them to be less sensitive to this irrelevant variability than they are to the relevant variability. This is known as invariant representation learning, an idea that has been around for decades \citep{fukushima1980neocognitron}.

The distortions represent a source of variability which does not change the class. Swaps, i.e. subtracting off the image and adding in another image, are important. Additive gaussian noise serves as a control in this experiment. The experiments probe responses to infinitesimal perturbations, which can be related to the subsequent theoretical analysis in Sections \ref{sec:LocallyLinear} and \ref{sec:Cancellation}. Finite perturbations give qualitatively similar results (data not shown).

\subsection{Learning sparse codes}
Given a dataset $\{\mathbf{x}(t) \in \mathbb{R}^m: t=1,2,\hdots,T\}$, a popular objective used to learn the dictionary $\mathbf{D} \in \mathbb{R}^{m\times n}$  and representations $\{\mathbf{r}(t) \in \mathbb{R}^m: t=1,2,\hdots,T\}$ was introduced by \cite{olshausen1996emergence}:
\begin{equation}
\begin{split}
    & \min_{\mathbf{D},\mathbf{r}(t)} \; \sum_{t=1}^T \frac{1}{2} \left\Vert \mathbf{x}(t) - \mathbf{D} \mathbf{r}(t) \right\Vert^2 + \lambda \left\Vert \mathbf{r}(t) \right\Vert_1 \text{  such that  } \left\Vert \mathbf{d}_i \right\Vert=1 \; \forall \; i 
\end{split}
\label{eqn:objective}
\end{equation}
In this paper we use the notation $\Vert \mathbf{u} \Vert$ indicates the $l_2$ norm of the vector $\mathbf{u}$. To indicate the $l_1$ norm we write $\Vert \mathbf{u} \Vert_1$. The hyperparameter $\lambda \geq 0$ controls the degree of sparseness. The dictionary elements are normalized to be unit vectors, $\left\Vert \mathbf{d}_i\right\Vert=1$, which prevents the representations from driving to zero to minimize their $l_1$ norm. 

We minimized Equation \ref{eqn:objective} for the 50,000 handwritten digits from the training split of the MNIST dataset.  Both the image and representation dimensionality are $784$. We vary $\lambda \in \{0.03,0.10,0.30,0.90,3.0\}$. 

The minimization is performed with an alternating gradient descent-like algorithm. Both the dictionary and representations are initialized from a unit normal distribution. For every training iteration we apply an ISTA update to the representations, and a gradient step to the dictionary, followed by normalization of every dictionary element $\mathbf{d}_i \leftarrow \mathbf{d}_i / \Vert \mathbf{d}_i \Vert$. The ISTA updates are explained in Eqs. 1.4,1.4 of \citet{beck2009fista}.

Separate learning rates for both $\mathbf{D}$ and $\mathbf{r}$ are chosen automatically and are updated at each step. Specifically, if the update decreased the loss, the update is accepted and the learning rate is multiplied by $1.1$. If the loss increased, the update is rejected and the learning rate is multiplied by $0.5$.

We then freeze the dictionary and continue to run ISTA updates (up to 200k iterations) until the representations are sufficiently close to optimal (defined in the Appendix) given the fixed dictionary. Reconstructions are highly accurate, and closely resemble the original images.

\subsection{Sensitivity of sparse codes}
\begin{figure}
    \centering
    \includegraphics[width=\linewidth]{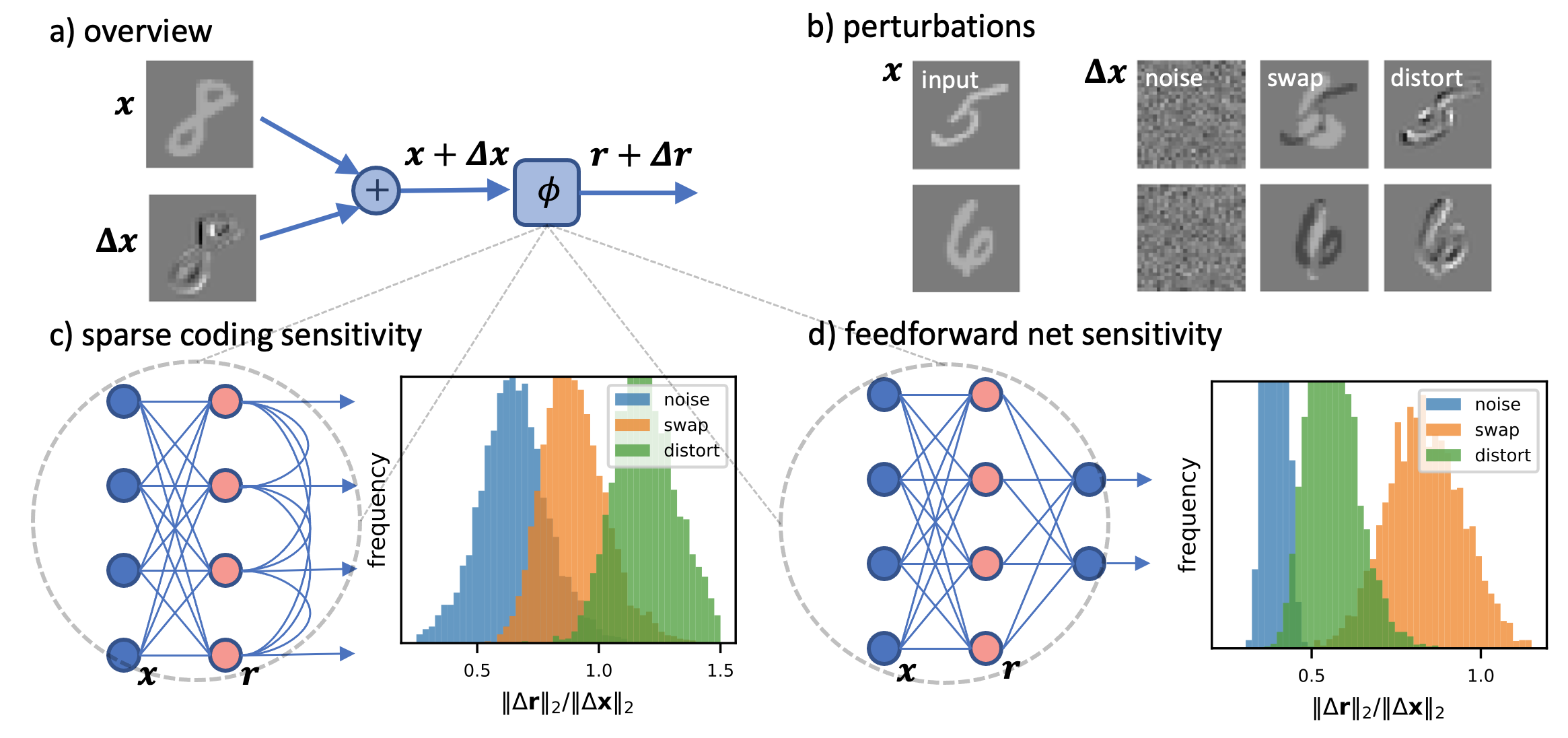}
    \caption{Representations generated by sparse coding are more sensitive to image distortions than to swapping the image. The the opposite behavior when examining the hidden layer of a fully connected network trained with supervised learning to discriminate digits. The sensitivity is measured by the norm of the directional derivative $\Vert \Delta \mathbf{r} \Vert / \Vert \Delta \mathbf{x} \Vert$ where $\Delta \mathbf{x}$ is an infinitesimal perturbation. Both representations are relatively less sensitive to noise than either warps or swaps.}
    \label{fig:input-perturbation}
\end{figure}

If the image $\mathbf{x}$ is perturbed by $\Delta\mathbf{x}$, its representation $\mathbf{r}$ changes by $\Delta\mathbf{r}$ (Figure \ref{fig:input-perturbation}a). We quantify sensitivity by the norm of the directional derivative $\Vert \Delta \mathbf{r}\Vert / \Vert \Delta \mathbf{x}\Vert$, to three types of infinitesimal perturbations $\Delta \mathbf{x}$: adding gaussian noise, swapping to another image, and distorting the image (Figure \ref{fig:input-perturbation}b). To compute the directional derivative $\Vert \Delta \mathbf{r}\Vert / \Vert \Delta \mathbf{x}\Vert$, we compute the Jacobian matrix $d\mathbf{r}/d\mathbf{x}$ (see Section \ref{sec:Cancellation} for equation). We then compute $\Vert d\mathbf{r}/d\mathbf{x} \cdot \Delta \mathbf{x}\Vert / \Vert \Delta \mathbf{x} \Vert$. This is repeated for 4000 samples and we show the distribution of the directional derivatives in Figure \ref{fig:input-perturbation}c.

For noise perturbations we set $\Delta \mathbf{x} = \mathcal{N}(0,I)$. For swap perturbations, we sample another image $\mathbf{x}_{swapped}$ and set $\Delta \mathbf{x} = \mathbf{x}_{swapped}-\mathbf{x}$. For distortion perturbations, we use the \textit{elasticdeform} augmentation module \citep{tulder2021elastic} which elastically warps the images as described in \cite{ronneberger2015unet}. For this experiment, a $5\times 5$ grid is generated with displacements sampled from a Gaussian distribution with $1/4$ pixel standard deviation. A $28 \times 28$ displacement field is then generated by upsampling with bicubic interpolation and the image is distorted with this displacement field. The distortions are small but observable if one carefully compares two different distortions of the same image. 

The sparse codes turn out to be most sensitive to distortions, followed by swaps, and finally noise (Figure \ref{fig:input-perturbation}c). Intuitively the noise invariance property makes sense: the network does not model noise, and this result suggests the network is at least just partially ignoring the noise. This is likely related to the well-known image denoising properties of sparse codes \citep{elad2006denoising}. The important behavior here is the greater sensitivity to distortions than swaps. This behavior is opposite that of a digit classifier, which should be less sensitive to distortions than swaps. (Swaps change the image to another randomly drawn image, usually that of another class altogether.) Therefore, the sparse codes seem to be making the classification task harder.

We note that the directional derivative $\Vert \Delta \mathbf{r}\Vert/ \Vert \Delta \mathbf{x} \Vert$ is normalized, so that the differences between the histograms in Figure \ref{fig:input-perturbation}c do not arise from differences in the size of perturbations $\Delta \mathbf{x}$. It is instructive to consider the baseline case where the image pixels themselves are used as the representation, $\mathbf{r}=\mathbf{x}$. In this case, $\Vert \Delta \mathbf{r} \Vert / \Vert \Delta \mathbf{x} \Vert=1$ for all kinds of perturbations (distortion, swap, and noise). In other words, the image itself is equally sensitive to all kinds of perturbations. 

In summary, we have found that sparse codes are more sensitive to distortions than to swaps. This sensitivity emerges in the sparse code and is not inherited from the image, which is equally sensitive to all kinds of perturbations. In other words, representing an image by a sparse code appears to take a step away from invariant object recognition, rather than a step towards it. The Appendix repeats the experiment of Fig. \ref{fig:input-perturbation} for multiple values of the sparsity parameter $\lambda$, and shows that greater sensitivity to distortions than swaps is a robust finding.

\subsection{Comparison with hidden layer of supervised feedforward net}
For comparison we also train a two layer fully connected network with supervised backpropagation to classify digits. The hidden layer is 784 dimensional, the same as our sparse representations, and the output is 10 dimensional. The hidden layer uses ReLU nonlinearity. We optimize the cross-entropy loss with stochastic gradient descent using a batch size of 64. We do not augment the images so the images are exactly the same ones used to train the sparse coding model. The network is trained for 1000 steps with a learning rate of 0.1 then 1000 more steps with a learning rate of 0.01. It achieves 98.7\% training accuracy and 97.7\% validation accuracy. This not intended to be a state-of-the-art network, but is sufficient for our purpose of comparison.

We consider the hidden layer of this network as a representation, and quantify its sensitivity to the same perturbations as before (Figure \ref{fig:input-perturbation} d). This representation turns out to be less sensitive to distortions than to swaps. In this sense, the hidden layer representation is taking a step towards invariant object recognition. This may not be surprising, as the network was trained to classify digits. Still, the hidden layer serves as an informative contrast with the sparse codes.

\section{Locally linear representations}\label{sec:LocallyLinear}
To help explain the preceding empirical findings, we provide a locally linear analysis of sparse codes in the following. For a fixed dictionary $\mathbf{D}$, the representation $\mathbf{r}$ of an image $\mathbf{x}$ is defined by minimizing the objective in Eq. (\ref{eqn:objective}) with respect to $\mathbf{r}$. 
The solution satisfies
\begin{equation}\label{eqn:NeuralNet}
    r_i = S_\lambda\left(\mathbf{d}_i\cdot\mathbf{x} - \sum_{j, j\neq i}(\mathbf{D}^\top \mathbf{D})_{ij}r_j\right)
\end{equation}
We will assume that the argument of the shrinkage function
\begin{equation}
    S_\lambda(u) = 
    \begin{cases}
    u-\lambda,& u\geq \lambda\\
    0, & |u|\leq \lambda\\
    u+\lambda, & u\leq -\lambda
    \end{cases}
\end{equation}
is ``not exactly at threshold,'' i.e., not equal to $\pm \lambda$ for any $i$. Violations of this assumption occur only for nongeneric images. Equation (\ref{eqn:NeuralNet}) has the form of a neural net in which $\mathbf{x}$ is the input and $\mathbf{r}$ is the output. This is the opposite of the generative model, in which $\mathbf{r}$ is the input and $\mathbf{x}$ is the output. In the neural net, $\mathbf{d}_i\cdot\mathbf{x}$ is a linear filtering of the image, so we will use the terms ``filter'' and ``dictionary element'' interchangeably to refer to $\mathbf{d}_i$.

The nonzero elements of $\mathbf{r}$, the active representation $\mathbf{r}_+$, can be written as a locally linear function of $\mathbf{x}$
\begin{equation}
    \mathbf{r}_+ = (\mathbf{D}^{\top}_+ \mathbf{D}_+)^{-1} \left(\mathbf{D}^{\top}_+ \mathbf{x} - \lambda \mathbf{s}_+ \right)
\label{eqn:active_response}
\end{equation}
Here $\mathbf{D}_+$ is the active dictionary and $\mathbf{s}_+$ contains the signs of the elements in $\mathbf{r}_+$.

A sufficient condition for uniqueness of $\mathbf{r}$ is that no dictionary elements are duplicates or antipodal pairs. (This follows from Lemma 3 of \cite{tibshirani2013uniqueness}, assuming that all dictionary elements are unit vectors.) The network used for Figure \ref{fig:input-perturbation} satisfies this condition. Another sufficient (and almost necessary) condition for uniqueness is that $\mathbf{D}_+$ should be nonsingular \citep{tibshirani2013uniqueness}, in which case the matrix inverse in Eq. (\ref{eqn:active_response}) is well-defined. 

\section{Non-orthogonality of filters can lead to sensitivity}\label{sec:Cancellation}

This sensitivity to distortions arises from the non-orthogonality of the dictionary. We'll argue that this sensitivity can partially be understood by considering the response to image perturbations which align with the difference $\mathbf{d}_a - \mathbf{d}_b$ between overlapping filters. We'll show that the representations generated by lasso inference are more sensitive to this difference than to either filter individually. For MNIST digits, the difference between the most overlapping pairs of filters often more closely resembles a spatial derivative of a filter, than the filter itself.

\subsection{Representations are more sensitive to filter differences to the filters themselves}

We start by considering two simple perturbations: one in the direction of an active filter $\Delta\mathbf{x}_1 = \epsilon \mathbf{d}_1$ and one in the direction of the difference between active filters $\Delta\mathbf{x}_{1,2} = \epsilon[\mathbf{d}_1-\mathbf{d}_2]$. Using Eq.\ (\ref{eqn:active_response}), it can be shown that the responses to these perturbations are $\Delta\mathbf{r}_1=\{+\epsilon,0,0,\hdots,0\}$ and $\Delta\mathbf{r}_{1,2}= \{+\epsilon,-\epsilon,0,\hdots,0\}$ for sufficiently small $\epsilon$, given the ``not exactly at threshold'' assumption. So the directional derivatives for the perturbations are
\begin{equation}
     \frac{\Vert \Delta \mathbf{r}_1\Vert}{\Vert \Delta \mathbf{x}_1 \Vert} = 1 \;\;\;\;\;\; \frac{\Vert \Delta \mathbf{r}_{1,2}\Vert}{\Vert \Delta \mathbf{x}_{1,2} \Vert} = \frac{1}{\sqrt{1-\mathbf{d}_1 \cdot \mathbf{d}_2}}
     \label{eqn:pairgain}
\end{equation}
using the fact that the filters are unit vectors. According to the second equation, $\mathbf{r}$ becomes more sensitive to difference perturbations as the overlap $\mathbf{d}_1 \cdot \mathbf{d}_2$ increases. The sensitivity diverges as the overlap approaches 1. When the overlap is exactly one, there is a continuum of minima for Eq. (\ref{eqn:objective}) and the representation is not unique. In this regard, the high sensitivity to difference perturbations can be regarded as a soft version of the non-uniqueness seen when there are two identical filters.

\subsection{Filter differences resemble spatial derivatives for the MNIST experiment}
\begin{figure}
    \centering
    \includegraphics[width=\linewidth]{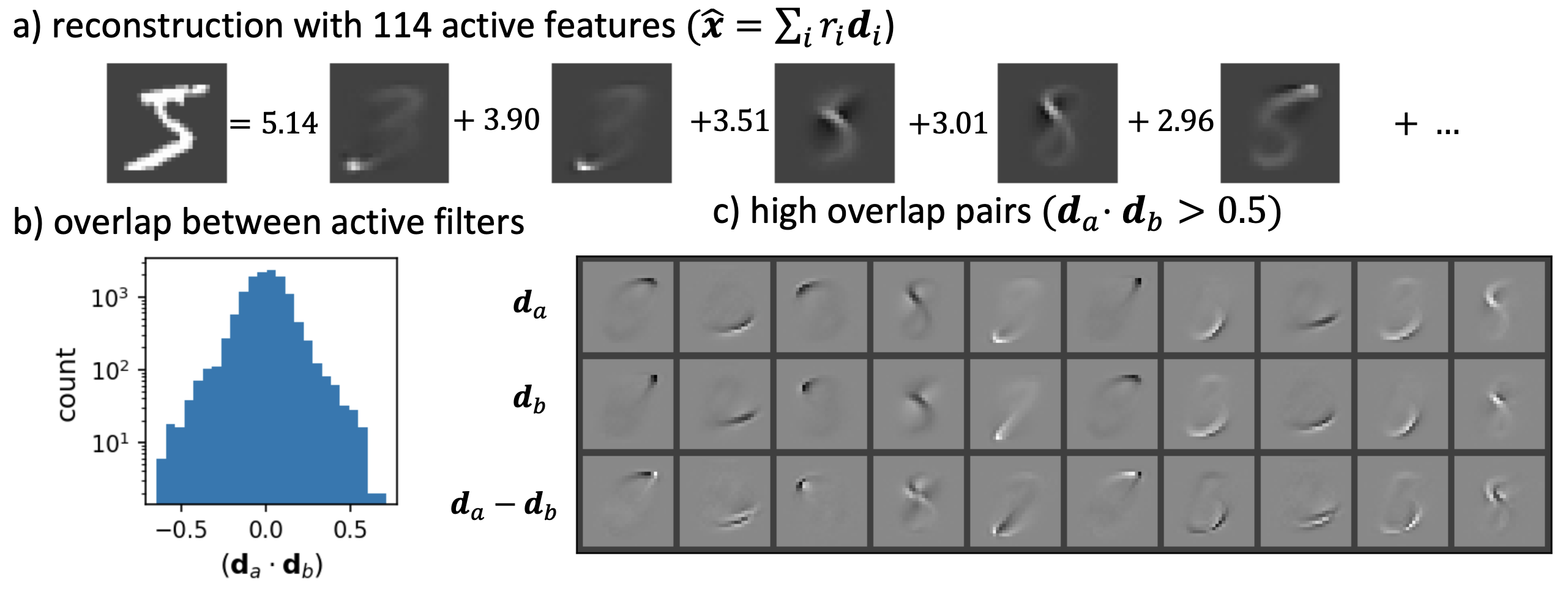}
    \caption{Sparse codes learned from MNIST lead to non-orthogonal filters. In (a) we show the reconstruction of the first MNIST digit and 5 of the 122 filters used in the reconstruction. In (b) we show the overlap distribution between all pairs of the 122 active filters. In (c) we show 10 pairs of filters with high overlap and their differences. The differences often more closely resemble spatial derivatives of the filters. The representations are more sensitive to image perturbations which look like these differences than perturbations which look like the filters themselves.}
    \label{fig:differences}
\end{figure}
Given Eq. (\ref{eqn:pairgain}), it is natural to ask whether in practice there exist filter pairs with high overlap. Figure \ref{fig:differences} returns to MNIST, and shows pairs of active filters with overlap greater than 0.5. The filters resemble parts (or negative parts) of digits. The differences between filters resemble distortions of the filters. Notably most filters have a significant mean value. For the 10 filter pairs in Fig. \ref{fig:differences}c, the average of the normalized means $ \frac{1}{10} \sum_{i=1}^{10} \sum_m d^a_{i,m} / \sum_m |d^a_{i,m}|$ is 0.66. The average of the mean of the difference $ \frac{1}{10} \sum_{i=1}^{10} \sum_m (d^a_{i,m}-d^b_{i,m}) / \sum_m |d^a_{i,m}-d^b_{i,m}|$ is only 0.09. This suggests that adding the filter difference is more like a transport/displacement of some of the pixels, compared to adding one of the filters which is more like an addition or removal of pixels.

\subsection{Why are the most similar filters distorted versions of one another?}

The observation that the most similar filters appear as distorted versions of one another is consistent with the predictions made in the sparse manifold transform framework \citep{chen2018smt}. Specifically it is postulated that the dictionary learned by sparse coding has ``has an organization that is a discrete sampling of a low-dimensional, smooth manifold'' \citep{chen2018smt}. Additionally it is postulated that the representations and dictionary act as a set of steerable filter which act to approximate continuous variations by. The idea is that in general, an oriented edge can show up at any continuous position in the image and will never perfectly align with an oriented edge in the dictionary. Weighted averages of dictionary elements can be used to approximate the continuous position of this edge.

The novelty of our work is to consider the implications of steerability when the filters being used to steer are non-orthogonal. The key theoretical insight is that when these filters have some positive overlap (i.e. they are discretely but sufficiently close along a manifold) the representations are extremely sensitive to small displacements of the image along the manifold.

\section{High sensitivity due to a highly cancelling combination of active filters}
In the previous section we showed that sparse coding can be sensitive to an image perturbation that is a linear combination of two filters that overlap, or equivalently tend to cancel each other.
We now show that the sensitivity can be even higher if there exists a linear combination of more than two filters with more cancellation. 

\subsection{Singular values and gain of amplification}
From the locally linear Eq. (\ref{eqn:active_response}), we can write the ``active Jacobian'' of the $\mathbf{x}\to \mathbf{r}$ map as the $k\times m$ matrix
\begin{equation}
    \mathbf{J}_+ := \frac{d\mathbf{r}_+}{d\mathbf{x}} = (\mathbf{D}^{\top}_+ \mathbf{D}_+)^{-1} \mathbf{D}^{\top}_+
    \label{eqn:jacobian}
\end{equation}
The $n-k$ rows of the Jacobian that are outside the active set have been omitted because they vanish. 
We will also refer to the active Jacobian as the ``gain matrix,'' because it helps us conceptualize the linear response as amplification by different gains for different directions in the image space.

The largest singular value of the gain matrix is the reciprocal of the smallest singular value of the active dictionary $\mathbf{D}_+$
Therefore, the representation $\mathbf{r}$ has the potential to be sensitive to changes in the image $\mathbf{x}$ if the active dictionary has a small singular value. We can write the smallest singular value $\sigma^\ast$ in variational form as
\begin{equation}
    \sigma^\ast = \min_{\mathbf{v}} \Vert \mathbf{D}_+ \mathbf{v} \Vert \text{ such that } \Vert \mathbf{v} \Vert=1
    \label{eq:MaximumCancellation}
\end{equation}
The solution $\mathbf{v}^\ast$ is a vector in the representation space, and can be interpreted as the coefficients of a linear combination of dictionary elements with maximum cancellation. The corresponding vector $\mathbf{u}^\ast \propto D_+\mathbf{v}^\ast$ in the image space is the direction that exhibits amplification with maximum gain $1/\sigma^\ast$. Both $\mathbf{v}^\ast$ and $\mathbf{u}^\ast$ are columns of the orthogonal matrices in the singular value decomposition (SVD).

\subsection{Image distortions align with high gain directions}
\begin{figure}
    \centering
    \includegraphics[width=\linewidth]{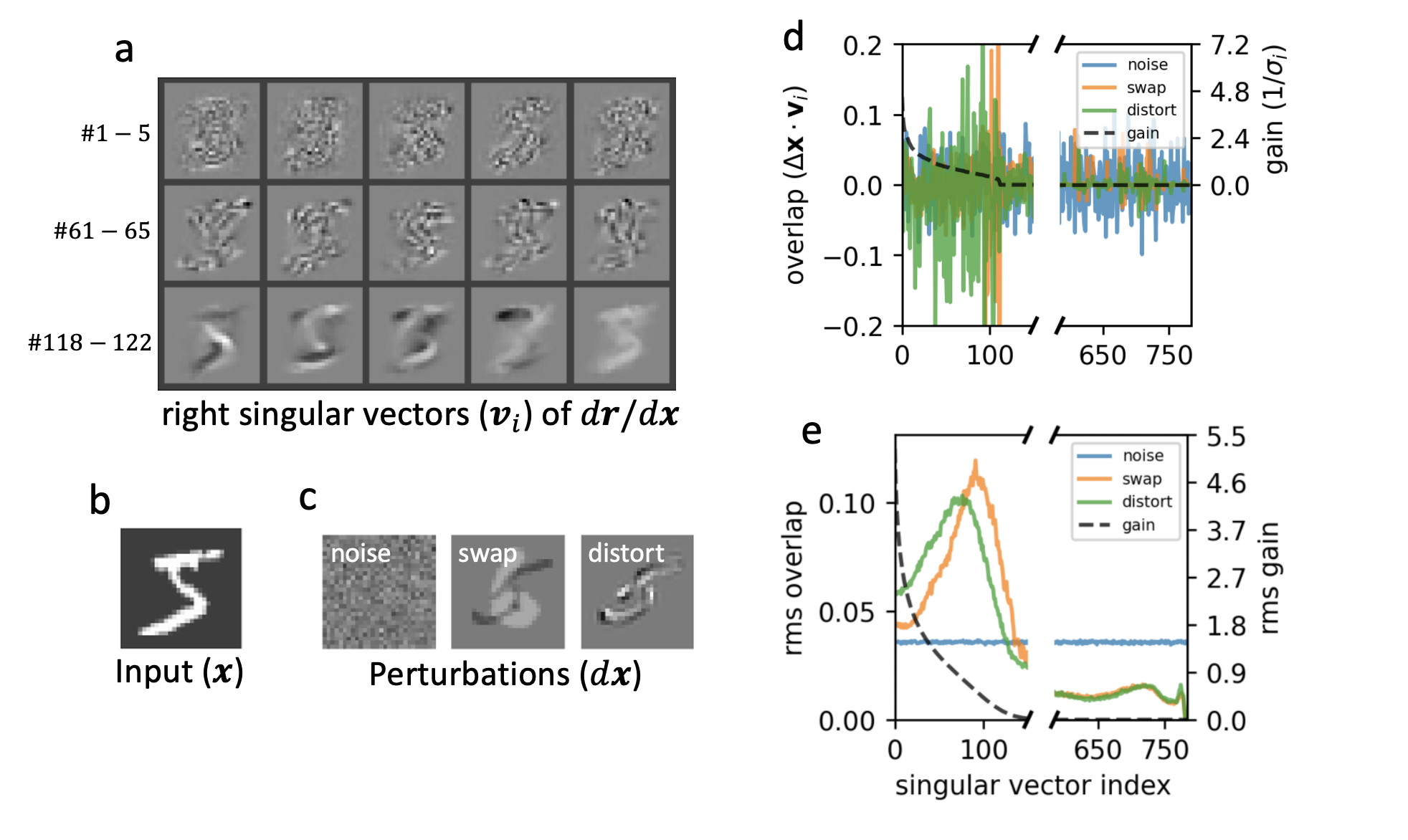}
    \caption{Singular value decomposition of the active dictionary, $\mathbf{D}_+ = \mathbf{U} \mathbf{\Sigma} \mathbf{V}^{\top}$, for the MNIST digit shown in b.  In (a) we show singular vectors in the image space (columns of $\mathbf{V}$), displayed as $28\times 28$ images in order of decreasing gain $1/\sigma$, where $\sigma$ denotes singular value. The vector with highest gain (top left) appears to be a high frequency noisy image, and the vector with lowest gain (bottom right) appears to be a blurred version of the original image. Only the top 122 singular vectors with nonzero gain are shown; they span the same subspace as the active dictionary. In (c) we show examples of three types of input perturbations. In (d) we plot the dot product between the singular vectors and the three input perturbations. We also plot the gain (1/$\sigma_i$) for each of these singular values. The gains vary by $20\times$ from the largest to the smallest nonzero gain. The perturbations of Fig. \ref{fig:input-perturbation} can be expressed as linear combinations of the singular vectors (colored lines). (e) Root-mean-squared overlap (``amplitude spectrum'') computed with the same 4000 perturbations and digits used to generate Fig. \ref{fig:jacobian}c. From noise to swap to distortions, the amplitude spectrum shifts towards the high gain vectors, explaining why the sensitivity of the representation increases.}
    \label{fig:jacobian}
\end{figure}

Returning to our MNIST experiment, we can find the image space direction of ``maximum cancellation'' by computing the SVD of the active dictionary or active Jacobian. 
The results are displayed in Fig. \ref{fig:jacobian}a for the first MNIST digit. 
The smallest singular value of the active dictionary is 0.158, corresponding to a maximum gain of 6.33. 

For comparison, achieving the same gain with a filter pair as in Eq. (\ref{eqn:pairgain}) would require overlap of 0.975. In fact, the maximum overlap between filters in the active dictionary is about 0.6 (see Fig. \ref{fig:differences}b), which amounts to a gain of only about 1.6 according to Eq. (\ref{eqn:pairgain}). The noisy appearance of the corresponding singular vector is consistent with the idea of maximum cancellation between active dictionary elements as in Eq. (\ref{eq:MaximumCancellation}).

We define the power spectrum of an image perturbation as the square of its overlaps on the singular vectors in the image space. The power spectrum always sums to one, assuming that the image perturbation is a unit vector. However, the distribution of power over the singular vectors differs for the different kinds of perturbations that were studied in Fig. \ref{fig:input-perturbation}. A noise perturbation has equal power on average over all singular vectors (Fig. \ref{fig:jacobian}c). That means most of the power of a noise perturbation is in the numerous singular vectors with zero gain, consistent with the experimental observation in Fig. \ref{fig:input-perturbation} that sparse coding is robust to noise perturbations. 

In contrast, both swaps and distortions have relatively little power in the singular vectors with zero gain, and more power in the singular vectors with nonzero gain. Furthermore, the power spectrum for distortions is shifted to the left, towards the singular vectors with higher gain. This is consistent with the experimental observation that sparse coding is more sensitive to distortions than swaps. Subjectively, the singular vectors with lowest nonzero gain do have more resemblance to swaps than distortions (Fig. \ref{fig:jacobian}), consistent with the difference between swaps and distortions in the average power spectrum. 

\subsection{Violation of Restricted Isometry Property}
In the field of compressed sensing, the dictionary $\mathbf{D}$ is often designed to satisfy the Restricted Isometry Property (RIP) \citep{candes2008restricted}, meaning that the norms of $\mathbf{D}\mathbf{r}$ and $\mathbf{r}$ are approximately equal for sparse $\mathbf{r}$. In sparse coding, the dictionary is learned rather than designed. Figure \ref{fig:jacobian} shows that the singular values of the active dictionary $\mathbf{D}_+$ can span a large range, so that the RIP is far from being satisfied.

\section{Supervised evaluation of sparse representations}
\begin{figure}
    \centering
    \includegraphics[width=\linewidth]{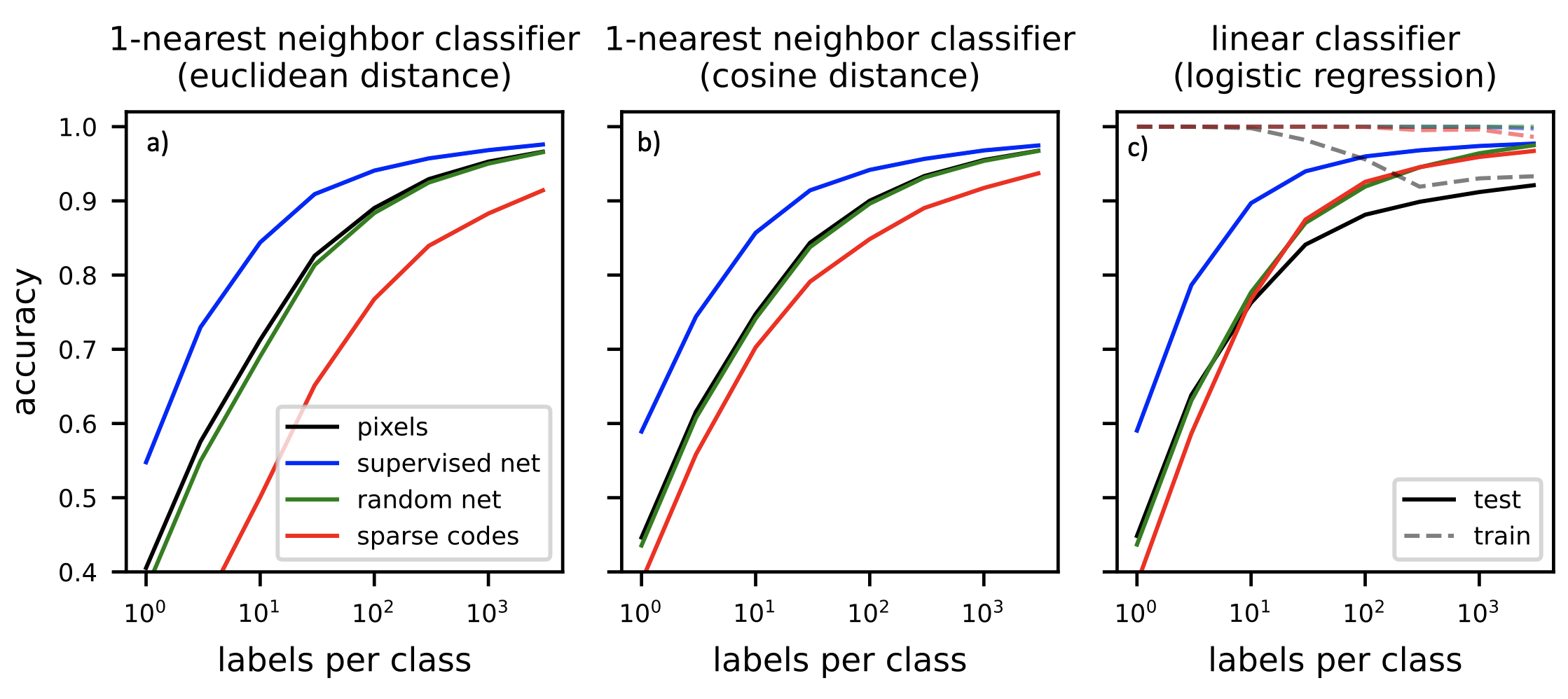}
    \caption{Supervised evaluation of four representations: pixels, sparse coding with $\lambda=0.3$, the hidden layer of a two-layer net trained via backpropagation to recognize digits, and a wide network with two layers of random weights (Eq. \ref{eqn:RandomNet}). For nearest neighbor classification (Euclidean or cosine distance), pixels and the random net yield nearly identical performance. The sparse representation is uniformly worse than pixels, and the hidden layer of the trained net is uniformly better. For linear classification, the sparse representation is worse than pixels with few labels and better than pixels with many labels. However, the sparse representation never outperforms the random net. In all cases, the hidden layer of the backprop-trained net is uniformly better than all other representations. }
    \label{fig:supervised}
\end{figure}

While generative models have many applications, here we are specifically interested in using the learned representations for classification. We evaluate four different sets of representations using nearest neighbor and linear classifiers. The representations we evaluate are given by 1) raw pixels $(\mathbf{x})$ 2) sparse coding $(\mathbf{r})$ with $\lambda=0.3$ 3) the hidden layer of a 2 layer net trained to recognize digits $(\mathbf{h})$ and 4) the 2nd layer of a wide network with random weights $(\mathbf{z})$. The first 3 are described in section 2. We describe the random net representations below. The results for $\lambda=0.03,0.1,0.9,3.0\}$ are similar and shown in the Appendix.

\subsection{Random feature baseline}
We evaluate our classifiers on a set of representations generated by a random 2-layer feedforward network with rectifying nonlinearity. Specifically, we create two matrices $\mathbf{W}_1$ of size $7840 \times 784$ and $\mathbf{W}_2$ of size $7840 \times 7840$ by sampling all elements iid from a unit normal distribution $W_{ij} \in \mathcal{N}(0,1)$. Defining $f(x)$ as the rectified linear unit (ReLU) nonlinearity, our representation $\mathbf{z}$ is generated by applying the 2 layer network to the image:
\begin{equation}\label{eqn:RandomNet}
    \mathbf{z}(\mathbf{x}) = f(\mathbf{W}_2 \cdot f(\mathbf{W}_1 \cdot \mathbf{x}))
\end{equation}

\subsection{Nearest Neighbor Classification}
We train a 1-nearest neighbor classifier on the representations. We compare euclidean and cosine distance. We train every configuration using $k\in\{1,3,10,30,100,300,1000,3000\}$ labels per class. We average the accuracy for every configuration over 5 random seeds. In Figure \ref{fig:supervised}, we show the results for both Euclidean and cosine distance. 

As expected the random features and pixel representations have nearly identical performance (the random representations do not change the ordering of the distances so long as the network is sufficiently wide). However, the sparse representations are markedly worse for all numbers of labels. This makes sense in light of our previous experiments: the representations are more sensitive to small translations than rescaling and swapping images. Intuitively, translations and deformations are precisely the sorts of variability that a representation should be insensitive to if a nearest neighbor classifier is to be trained on top of it.

Interestingly, using cosine distance instead of Euclidean distance seems to have the largest positive impact on the sparse representations, though they still lag behind the other representations for all numbers of labels. This suggests that the norm of the sparse representations may be relatively unstable compared to pixel representations and by using cosine distance, we ignore variations in the length of the representation vector.

It is easy to dismiss the results of this nearest neighbor classification because it is not a particularly ``smart'' classifier: it makes few assumptions about the data and in many practical datasets it requires too many labels to achieve acceptably high generalization performance. However we believe this strong reduction in generalization performance shows a very real and important property of the sparse representations.

First, the criticism of the classifier as being too weak is not as relevant for MNIST digits: the classifier achieves $97.5\%$ validation accuracy when trained on pixels using Euclidean distance. Essentially this classifier works well on pixels, and the sparse representations "break" this classifier. This reduction in performance can partially be mitigated by normalizing the representations (i.e. using cosine instead of Euclidean distance) although not completely.

Second, this classifier provides a somewhat direct evaluation of the geometry of representations. This classifier gives us a formal method for measuring the degree of invariance of a set of representations. If the representations are in fact invariant to transformations of the image which are not important for recognition, while being sensitive to transformations which are important, then a nearest neighbor classifier will perform quite well. If the classifier does worse, this suggests the representations are "less invariant" than the original image.

Third, this classifier is in fact used to evaluate modern successful self-supervised learning systems \citep{caron2020swav}. Pretraining networks to be distortion invariant often results in substantial improvements in nearest neighbor classification even on much more challenging datasets like IMAGENET.

Fourth, the issue of training accuracy is not relevant for a nearest neighbor classifier. This makes the results simpler to interpret than most other classifiers. Even for a linear classifier, understanding generalization can be complicated in part because the representation may be improving linear separability, an effect easily achieved with random nonlinear features as we'll show in the next section.

\subsection{Linear Classification}
We train a standard linear classifier with logistic regression on the representations. We use a variable number of labels and 5 seeds for configuration like we did with for nearest neighbor classifiers. Additionally, we have an another parameter $\lambda_w$ controlling the strength of weight decay $\Vert \mathbf{w} \Vert$. For every configuration, we find the $\lambda_w \in \{1e-5,1e-4,1e-3,1e-2,1e-1,1\}$ which yields the highest test accuracy.

With more than 10 labels per class, both the random-feature and sparse representations improve the train and test accuracy of the linear classifier over raw pixels. With fewer than 10 labels per class all 3 representations perform comparably in train accuracy since they all perfectly fit the training set. This time, the random-feature and pixel representations perform similarly and the sparse representations perform worse. This reduced performance is not substantial, but it is robust across random seeds.

This experiment suggests a pessimistic explanation for the seemingly encouraging observation that the sparse representations improve generalization performance of a linear classifier. This experiment suggests the sparse representations are merely more linearly separable than pixels and the linear classifier is able to overcome the instabilities introduced by the sparse coding representation if it has enough labels. When you don't have enough labels, the linear classifier is outperformed by the raw pixels, which we argue is because the representations are so sensitive to distortions. The random-feature representations similarly improve the linear separability but they never hurt the linear classifier. And for the nearest neighbor classifier, the sparse representations are always worse than for pixels and random-feature representations. The Appendix shows that the above findings regarding classification are robust to the specific choice of the sparsity hyperparameter $\lambda$.

\section{Discussion}
A priori, there is no reason to expect that optimizing for reconstruction error will produce sparse representations that help with the downstream task of object recognition. We have argued that by some measures sparse coding produces representations which are less suited for subsequent recognition than the original pixel representations. We argued empirically that the representations can be particularly sensitive to image distortions that do not change the object class. 

A linear classifier trained on top of sparse codes has been shown to exhibit more ``adversarial robustness'' than a backprop-trained feedforward net \citep{paiton2020robustness}. This robustness seems related to our work in two ways. First, adversarial perturbations of the feedforward net resemble noise, and sparse codes are robust to noise (Fig. \ref{fig:input-perturbation}). Second, the robustness is claimed for the output of the linear classifier, not the sparse code itself.

We also train a linear classifier on top of sparse codes, and its output can be robust to distortions if trained with many class labels. However we observe worse than pixel performance with few labels, which we argue is an outcome of the distortion sensitivity of the underlying sparse code.

This sensitivity arises from nonorthogonality of the learned dictionary elements. A possible remedy is to add a term to the objective function that encourages the dictionary elements to be orthogonal \citep{sulam2020adversarial}, though this may be less effective as the dictionary size grows large.

It is interesting to contrast sparse coding with contemporary  self-supervised learning techniques \citep{dosovitskiy2014discriminative,chen2020simclr,he2020moco,caron2020swav,deny2021barlowtwins}. Networks are trained to yield similar representations for two distorted versions of the same image. Sparse codes, in contrast, are sensitive to image distortions.

Beyond sparse coding, there are other generative models such as VAEs \citep{kingma2014vae}, GANs \citep{goodfellow2014gan}, and autoregressive models \citep{oord2016pixelcnn}. These models generate beautifully, but have not seen the same level of success for object recognition as supervised training with lots of labels or contrastive pretraining. These generative models are more complex and harder to understand than sparse coding but we can speculate that perhaps these are other examples where reconstructing the original image is simply not an optimal pretraining task. This speculation might be testable via experiments similar to those in Figures \ref{fig:input-perturbation} and \ref{fig:jacobian}.

\section*{Acknowledgements}
We would like to thank L. And J. Jackel, Lawrence Saul, and Runzhe Yang for their helpful insights and discussions. We would also like to Dylan Paiton and Bruno Olshausen for their feedback which helped to both make the paper’s claims more precise and the experimental analysis more rigorous.

This research was supported by the Intelligence Advanced Research Projects Activity (IARPA) via Department of Interior/ Interior Business Center (DoI/IBC) contract number D16PC0005, NIH/NIMH RF1MH117815, RF1MH123400. The U.S. Government is authorized to reproduce and distribute reprints for Governmental purposes notwithstanding any copyright annotation thereon. Disclaimer: The views and conclusions contained herein are those of the authors and should not be interpreted as necessarily representing the official policies or endorsements, either expressed or implied, of IARPA, DoI/IBC, or the U.S. Government.

\section*{Appendix}
\subsection{Learned features for various $\lambda$}
\begin{figure}[h]
    \centering
    \includegraphics[width=\linewidth]{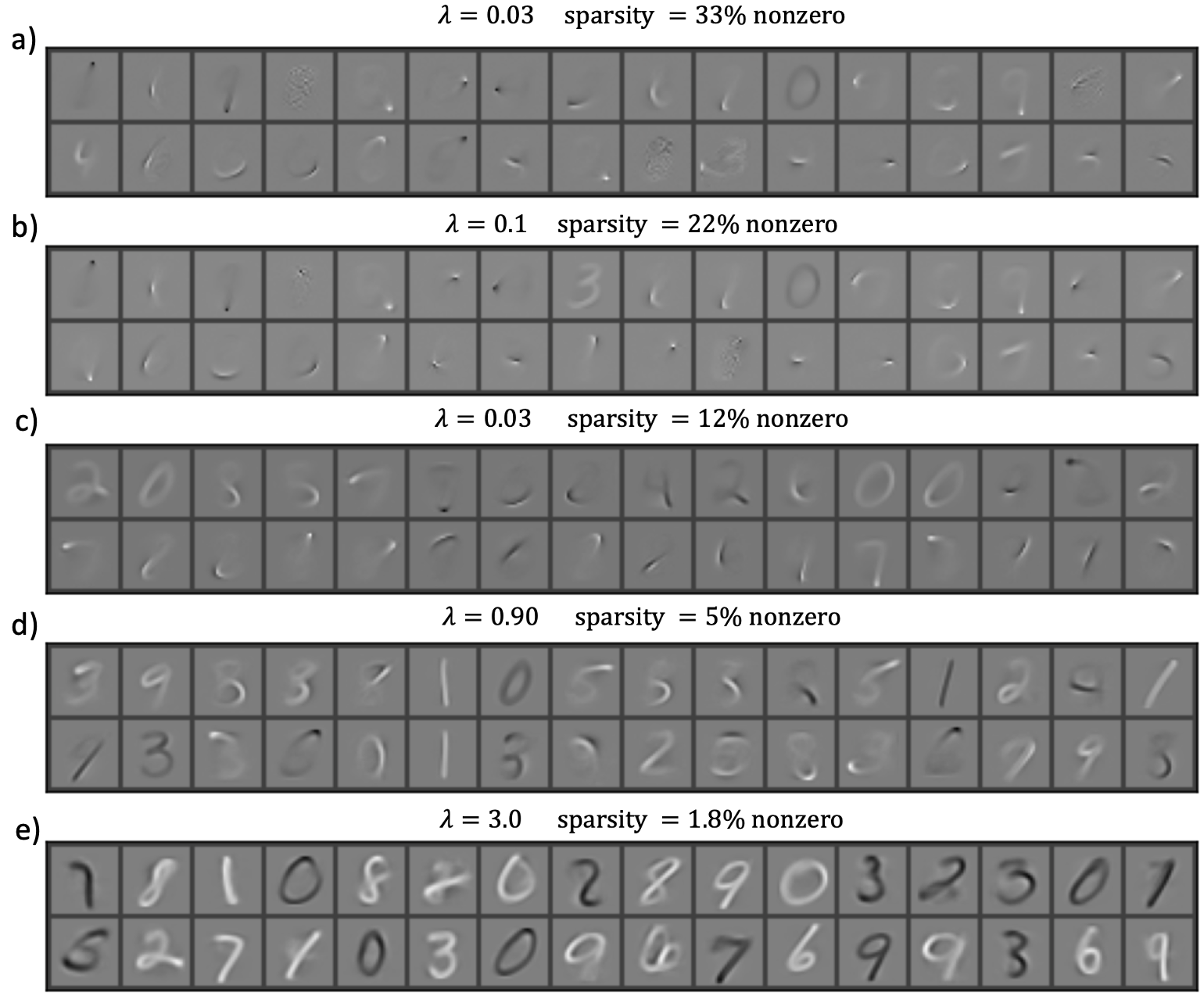}
    \caption{Elements of the dictionary learned for $\lambda \in \{0.03, 0.10, 0.30, 0.90, 3.0\}$. As $\lambda$ increases, the fraction of nonzero neurons decreases, and the dictionary elements move from resembling small spots and strokes ($\lambda =0.0$ up to whole digits $\lambda=3.0$)}
    \label{fig:features-all_lam}
\end{figure}

We train networks with $\lambda \in \{0.03,0.10, 0.30, 0.90, 3.0\}$, keeping the number of features the same in each case ($n=784$). We use the same alternating gradient-based optimization procedure described in Section 3.2. The only difference is we use more iterations for smaller $\lambda$ and fewer iterations for higher $\lambda$. Figure \ref{fig:features-all_lam} shows the learned dictionaries and the fractions of active neurons (over the training set) for the exact representations derived from each dictionary.

\subsection{Computing the representations for every image} At the end of the training process described in Section 3.2, we freeze the dictionary and continue to update the representations so we nearly have the exact unique representations for every image. As discussed in the main text, uniqueness of representations is guaranteed because the learned dictionary elements turn out to be unique (the largest cosine similarity between any pair is $0.9$). 

These representations are generated by running enough ISTA steps until Eq. (\ref{eqn:NeuralNet}) is satisfied to high numerical precision. For each image, we require:
\begin{equation}
\begin{split}
    & \text{ if } r_i = 0 \text{ then } | \mathbf{d}_i \cdot \mathbf{x} - \sum_j \mathbf{d}_i \cdot \mathbf{d}_j r_j | - \lambda < 10^{-5} \\
    & \left\Vert \mathbf{r}_+ - (\mathbf{D}^{\top}_+ \mathbf{D}_+)^{-1} (\mathbf{D}^{\top}_+ \mathbf{x} - \lambda \mathbf{s}_+) \right\Vert / \left\Vert \mathbf{r}_+ \right\Vert  < 10^{-4}
\end{split}
\label{eqn:convergence}
\end{equation}
where $\mathbf{r}_+$ is the active representation for a given image and $\mathbf{s}_+$ contains the signs of these elements $(s_+)_i = \text{sign}((r_+)_i)$. 

Finding the representations which achieve these conditions is time consuming. We start with single precision ISTA updates, and switch to double precision if the energy fails to decrease. The learning rate parameter is adjusted automatically as described in Section 3.2. We check the conditions Eq. \ref{eqn:convergence} for each image every 10,000 iterations.  Aapproximately 200,000 ISTA updates were required for convergence. In total it took approximately 12 GPU hours to compute all the representations for $\lambda=\{0.03,0.10,0.30,0.90,3.0\}$.

\subsection{Sensitivity histograms for various $\lambda$} 
\begin{figure}[h]
    \centering
    \includegraphics[width=\linewidth]{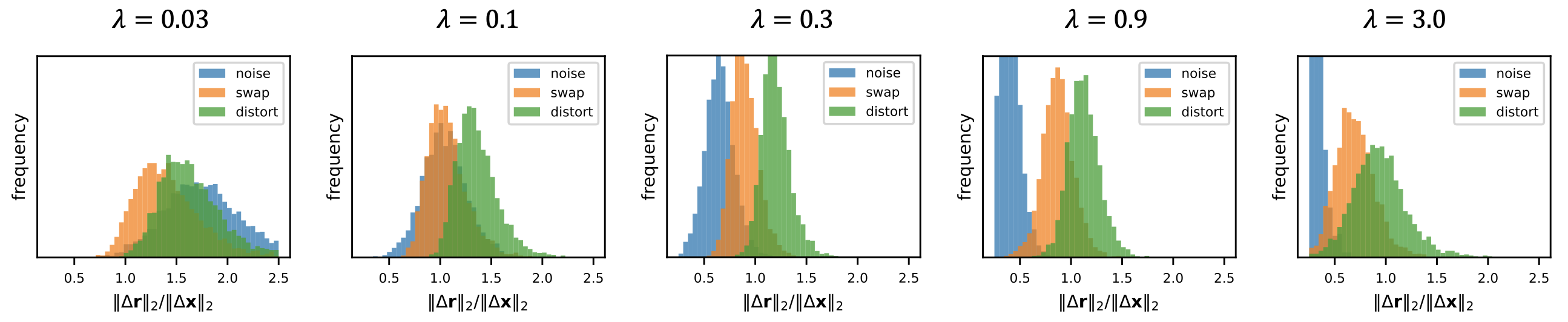}
    \caption{Sensitivity of the sparse representations to noise, swaps, distortion perturbations (examples are shown in Fig. 1). For every $\lambda$ we test, the sparse representations are more sensitive to distortions than to swaps. This difference seems most pronounced for the intermediate values of $\lambda = 0.1,0.3,0.9$. The noise sensitivity appears to be most strongly impacted by the sparsity level, and the representations are less noise sensitive as the sparsity level increases.}
    \label{fig:distorsions-all_lam}
\end{figure}

Figure \ref{fig:distorsions-all_lam} shows the sensitivity of the sparse codes to swaps, distortions, and noise. This histogram was shown for $\lambda=0.3$ in Figure 1c. We now show the histograms for $\lambda \in \{0.03,0.10,0.30,0.90,3.0\}$.

\clearpage

\subsection{Supervised evaluation for various $\lambda$}
\begin{figure}[h]
    \centering
    \includegraphics[width=\linewidth]{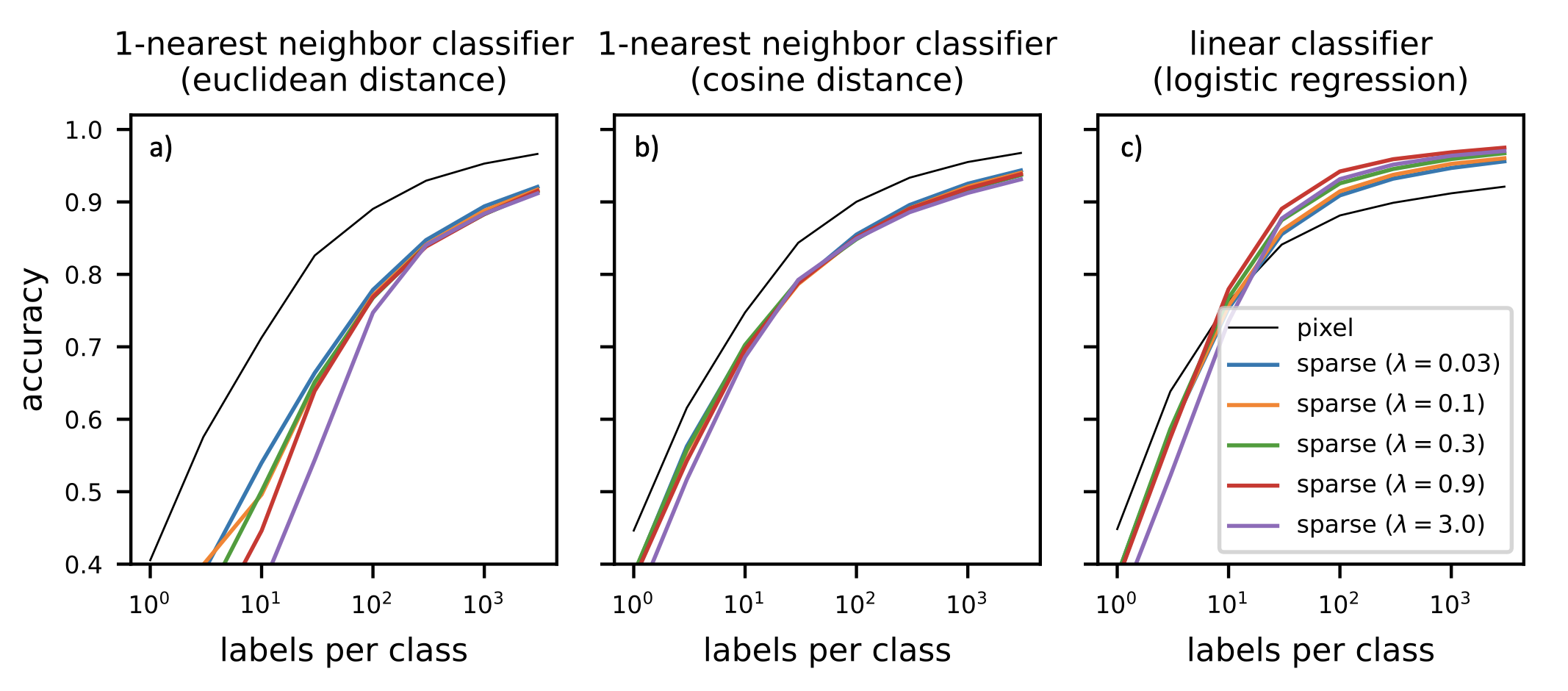}
    \caption{Supervised evaluation of the representations with various $\lambda$. The behavior of the representations is remarkably similar, given the difference in sparsity and learned features (Fig. \ref{fig:features-all_lam}) between the various $\lambda$. We do observe that an increase in sparsity is somewhat worse for both the nearest neighbor classifiers and the linear classifier when the number of training labels is low. }
    \label{fig:supervised-all_lam}
\end{figure}

Figure \ref{fig:supervised-all_lam} shows the supervised evaluation metrics for various $\lambda$.

\bibliographystyle{apalike}

\end{document}